\documentclass[letterpaper, 10 pt, conference]{format/ieeeconf}

\usepackage{amsmath}
\usepackage{amsfonts}
\usepackage{amssymb}

\usepackage{amsthm}
\usepackage[ruled,vlined]{algorithm2e}
\usepackage{mathtools}
\usepackage{tabularx}
\usepackage{graphicx}
\usepackage{subfigure}
\usepackage{enumerate}
\usepackage{float}
\usepackage{url}
\usepackage{verbatim}
\usepackage{stfloats}
\usepackage{bbding}
\usepackage{hyperref}
\usepackage{booktabs}
\usepackage{threeparttable}  
\usepackage{multirow}
\usepackage{pifont}  

\graphicspath{{figures/}}
\IEEEoverridecommandlockouts
\newcommand{\abs}[1]{\vert{#1}\vert}

\newcommand{\sref}[1]{Section \ref{#1}}

\makeatletter
\def\@fnsymbol#1{\ensuremath{\ifcase#1\or \dagger\or \ddagger\or
		\mathsection\or \mathparagraph\or \|\or **\or \dagger\dagger
		\or \ddagger\ddagger \else\@ctrerr\fi}}
\makeatother

\title{\textbf{FAST-Dynamic-Vision: \\ Detection and Tracking Dynamic Objects with Event and Depth Sensing}}
\author{Botao He$^{\dag \,}$\textsuperscript{3,4}, 
        Haojia Li$^{\dag \,}$\textsuperscript{3,5}, 
        Siyuan Wu\textsuperscript{3,6}, 
        Dong Wang\textsuperscript{1,3}, 
        \\ 
        Zhiwei Zhang\textsuperscript{1,3}, 
        Qianli Dong\textsuperscript{3,5}, 
        Chao Xu\textsuperscript{1,2,3}, 
        and Fei Gao\textsuperscript{1,2,3}
	\thanks{\textbf{${\dag}$ Equal contribution.}}        
    \thanks{1 State Key Laboratory of Industrial Control Technology, Institute of Cyber-Systems and Control, Zhejiang University, Hangzhou, 310027, China.} 
	\thanks{2 Huzhou Institute of Zhejiang University, Huzhou, 313000, China.}
	\thanks{3 Nation Engineering Research Center for Industrial Automation (Ningbo Institute), Ningbo, 315000, China}
    \thanks{4 School of Automation, Nanjing Institute of Technology, Nanjing, 211112, China.} 
	\thanks{5 Faculty of Robot Science and Engineering, Northeastern University, Shenyang, 110207, China.} 
	\thanks{6 Faculty of Electronic and Information Engineering, Xi’an Jiaotong University, Xi'an, 710049, China.}
	\thanks{Email:{\tt\small \{cxu, fgaoaa\}@zju.edu.cn}}}

\begin{document}

\maketitle
\thispagestyle{empty}
\pagestyle{empty}

\begin{abstract}

The development of aerial autonomy has enabled aerial robots to fly agilely in complex environments.
However, dodging fast-moving objects in flight remains a challenge, limiting the further application of unmanned aerial vehicles (UAVs). 
The bottleneck of solving this problem is the accurate perception of rapid dynamic objects.
Recently, event cameras have shown great potential in solving this problem.
This paper presents a complete perception system including ego-motion compensation, object detection, and trajectory prediction for fast-moving dynamic objects with low latency and high precision. 
Firstly, we propose an accurate ego-motion compensation algorithm by considering both rotational and translational motion for more robust object detection. 
Then, for dynamic object detection, an event camera-based efficient regression algorithm is designed.
Finally, we propose an optimization-based approach that asynchronously fuses event and depth cameras for trajectory prediction.
Extensive real-world experiments and benchmarks are performed to validate our framework.
Moreover, our code will be released to benefit related researches.

\end{abstract}

\section{Introduction}
\label{sec:Introduction}
Thanks to recent progress in aerial autonomy, UAVs have been able to fly agilely in complex environments such as mine exploration. 
Drones are able to perceive unknown environments and plan an exploration path autonomously. 
However, perception in dynamic environments, especially with high-speed objects, is still a challenging problem. 
For example, drones have difficulty dodging a rock falling head-on during the fast mine exploration.

For fast-moving object's avoidance, it's pivotal to track them and predict their future trajectories in a short latency.
Normally, this latency is hundreds of milliseconds for most perception methods: cameras need tens of milliseconds to expose and suffer from motion blur; besides, algorithms need a sequence of frames to predict a trajectory.
However, for objects with speed higher than 10 meters per second, such long latency leaves drones no time to escape. 
In order to reduce this latency, sensors with a higher temporal resolution are keenly demanded. 
Meanwhile, a real-time detection and tracking algorithm is also indispensable.

To fill this research gap, we adopt the event camera, an asynchronous motion-activated sensor providing a microsecond-level temporal resolution, for solving this problem.
In this work, a complete perception system integration for this sensor is also designed.
Firstly, we propose an ego-motion compensation algorithm to alleviate the noise. 
Then, for dynamic object detection, we develop a regression-based approach to find the region of interest (ROI).
Notably, this approach is more robust and less computational demanding compared to other clustering-based methods in \cite{falanga2020dynamic}.

Furthermore, a satisfactory solution should be capable of tracking the object in the 2D camera space and estimating its corresponding 3D trajectory~\cite{su2017catching}. 
To address the scale ambiguity issue, we further incorporate a depth camera to recover the scale of monocular sensing by joint optimization.
Afterward, combining event and depth observations, we present an accurate trajectory estimator which significantly increases the robustness and accuracy. 
Our algorithm successfully balances the tight onboard computational budget and trajectory accuracy.

We perform extensive quantitative and qualitative experiments in high dynamic scenarios to validate our object detection and trajectory estimation framework, which provide a solid foundation for fast-moving object avoidance.

\begin{figure}[t]
	\centering
	\includegraphics[width=1.0\linewidth]{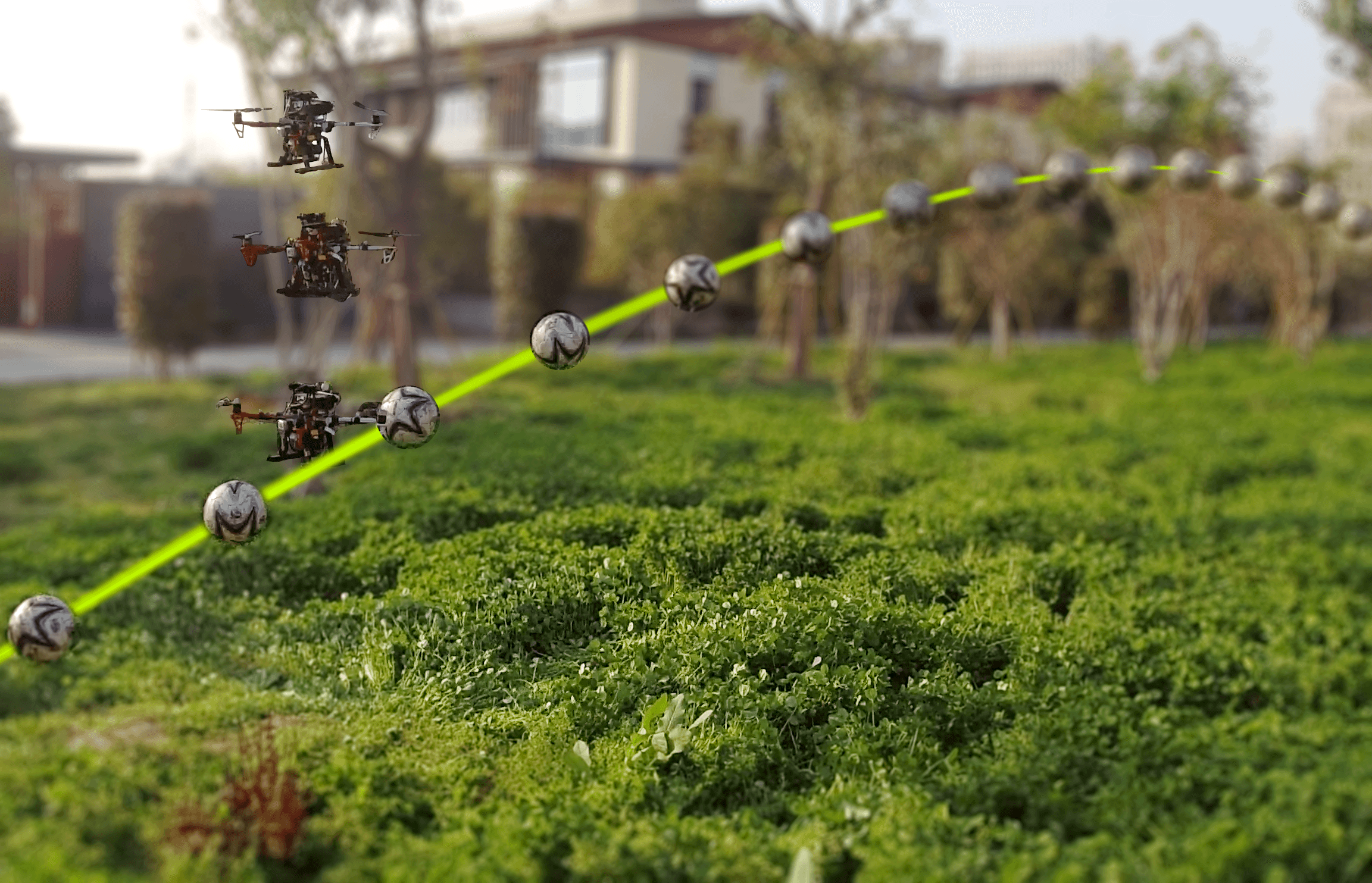}
	\caption{A composed image of real-world dodging experiment to validate our detection and tracking system. Please refer to our video submission for more details.}
	\label{fig:top}
\end{figure}

This paper highlights several features:

\begin{itemize}
	\item[1)] An advanced motion compensation method for event-detection balancing efficiency and accuracy.
	\item[2)] A 3D trajectory estimation approach that fuses event and depth information asynchronously.
	\item[3)] A complete system integration with open source \footnote{ Our code and video can be found at \url{https://github.com/ZJU-FAST-Lab/FAST-Dynamic-Vision}}.
\end{itemize}

\begin{figure*}[ht]
	\centering
	\includegraphics[width=\textwidth]{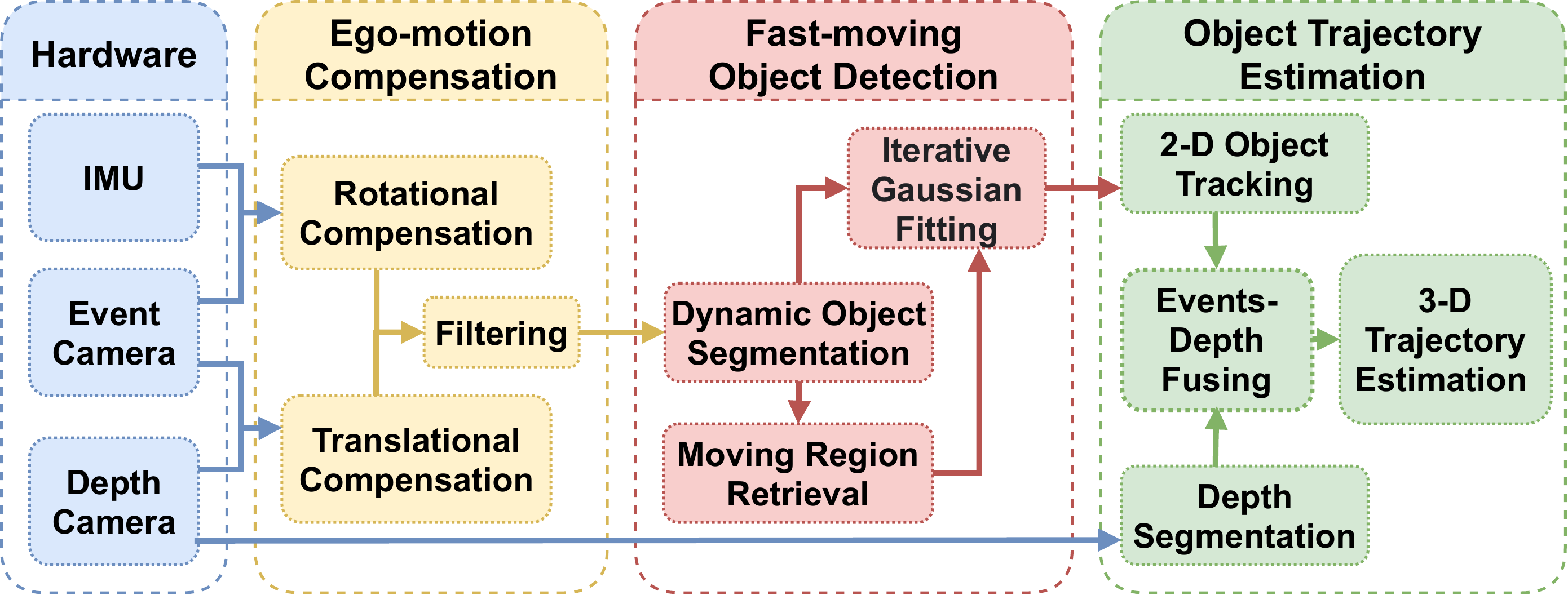}
	\caption{The overview of our detection and tracking system.}
	\label{fig:system_architecture}
\end{figure*}

\section{Related Work}
\label{sec:RelatedWork}

Due to its unique advantage of high temporal resolution and no motion blur, the event camera has attracted many researchers' interest \cite{gallegoEventbasedVisionSurvey2020} \cite{brosch2015eventbased}.
The first problem to solve is how to represent events.
Existing event representation algorithms can be divided into two main categories.
One category is classifying events of different objects into several clusters  \cite{zhu2017eventbased} \cite{Piatkowska2012Spatiotemporal} \cite{lagorce2015asynchronous} \cite{li2019robust} \cite{brandliELiSeDEventbasedLine2016}. 
This sort of method is intuitively built on event mechanisms, but it is sensitive to noisy events.
It also neglects time information which is vital for event-based detection. 
The other category is tracking features on time surface frames \cite{falanga2020dynamic} \cite{mitrokhin2018eventbased}  \cite{gallego2018unifying} \cite{zhou2018semi}, which is a 2D map only containing the latest event's timestamp while ignoring others triggered recently in each pixel. 
Specifically, some researchers
\cite{falanga2020dynamic} \cite{mitrokhin2018eventbased} introduces a mean-time image representation contains the average timestamp of the events.
This mean-time image is less computational demanding than other types, e.g., exponential time surfaces\cite{zhou2021eventbased}.
Furthermore, this representation is more suitable for object detection tasks: regions containing moving objects can be obtained by merely thresholding the mean-time image. 
Therefore, our detection method is based on it.

To remove background events generated by rotation and translation, ego-motion compensation is necessary for moving object detection and tracking.
Mitrokhin \textit{et al.} \cite{mitrokhin2018eventbased}  minimize error functions provided by spatial gradient of mean-time image to fit a parametric motion model;
Gallego \textit{et al.} \cite{gallego2018unifying} maximize a variance which represents local contrast, in other words, sharpness, on the compensated image.
Zhou \textit{et al.} \cite{zhou2020event} minimize an energy function. 
The optimization-based method is accurate. However, one drawback of this method is that its high computational cost introduces extra latency in the perception system \cite{falanga2020dynamic}, which would lead to potential failure in our object avoidance scene.
Falanga \textit{et al. }\cite{falanga2020dynamic} use IMU's angular velocity average to perform rotational ego-motion compensation. 
This method is less computationally demanding so that it can be applied for onboard flights while the accuracy is not guaranteed in forwarding flights.
Based on this method \cite{falanga2020dynamic}, we improve the motion compensation approach by fusing depth and IMU data to implement both rotational and translational ego-motion compensation.
Our method can enhance its accuracy and reliability without sacrificing computational efficiency.


For object tracking and trajectory estimation, our framework is inspired by the following studies.
Su \textit{et al.} \cite{su2017catching} fit a parabolic model to estimate the 3D trajectory of a flying object from noisy 2D observation.
This method requires plenty of observations due to the lack of depth information, which cannot meet the requirement for low-latency.
Falanga \textit{et al.} \cite{falanga2020dynamic} apply stereo event cameras for 3D position estimation.
However, this configuration does not guarantee accuracy and robustness because the high level of noise causes uncertainty in depth estimation.
To obtain more accurate 3D trajectories, we design a different configuration fusing event and depth sensor onboard.

\section{Overview}
\label{sec:Overview}
\subsection{System Architecture}
\label{Overview:SystemArchitecture}

The pipeline of our framework is illustrated in Fig \ref{fig:system_architecture}. 
There are three procedures in this framework: ego-motion compensation, object detection, and object trajectory estimation.
Firstly, we implement an advanced motion compensation algorithm fusing IMU and depth data to filter out background events generated by ego-motion, including rotation and translation during flight. 
The mean-time image can be generated by motion-compensated events. 
Each pixel value of this mean-time image is the average timestamp of corresponding events.
Following the motion compensation step, we detect and locate the region with the largest average timestamp in the mean-time image.
This region represents the area with the fastest speed on the image plane.
To obtain the region's bounding box, we introduce an iterative Gaussian fitting algorithm for the object detection step.
We also present a moving region retrieval to guarantee the bounding box we get is the most accurate one.
Next, the moving object's location is tracked with Kalman Filter on the 2D plane, and the object is segmented out on the depth map according to the detection result. 
Then, we optimize the trajectory of the object by minimizing reprojection residuals.
Finally, to validate our estimation, we design a scenario in which a UAV autonomously detect and avoid objects flying towards it.

The rest of this paper is organized as follows: 
\sref{Meth:EgoMotionCompensation} presents our advanced ego-motion compensation algorithms. 
Then we discuss object detection and tracking methods used in this framework (\sref{Meth:ObjDetection}).
In \sref{Meth:ObjTrajEst}, we perform our 3D trajectory estimator fusing event stream and depth information.
\sref{sec:Results} depicts our real flight experiment and compares our performance with others.

\subsection{Notation}
\label{Overview:Notation}

Let $ C \in \mathbb{R}^3 $ denote a set of events.
We use symbols $(x, y, t) \in C$ to denote an event triggered by an event camera. The symbol $x$, $y$ represents the event's coordinate on the image plane, $t$ denotes the timestamp of the event.

We represent $\xi_{i, j}$ as a set of motion-compensated events (see \ref{Meth:EgoMotionCompensation}) which are projected onto the same pixel $(i, j)$
\begin{equation}
	\xi_{i j}=\left\{\left\{x^{\prime}, y^{\prime}, t\right\}:\left\{x^{\prime}, y^{\prime}, 0\right\} \in C^{\prime}, i=x^{\prime}, j=y^{\prime}\right\} .
\end{equation}

Therefore, the event-count image pixel \cite{mitrokhin2018eventbased} can be denoted as $\mathcal{I}_{i,j}$ where
\begin{equation}
	\mathcal{I}_{i,j} = |\xi_{i,j}| .
\end{equation}

We also define the time-image as $\mathcal{T}$. 
Hence, pixel $(i, j)$ in the time-image represented as $\mathcal{T}_{i,j}$, can be expressed as the average timestamp of events triggered in this position, as follows: 
\begin{equation}
	\mathcal{T}_{i,j} = \frac{1}{\mathcal{I}_{i,j}} \sum t:t\in \xi_{i,j}  .
\end{equation}

We name the normalized time-image $\mathcal{T}$ as normalized mean-time image $\mathcal{N}$, which can be computed by the following equation \cite{falanga2020dynamic}
\begin{equation}
	\mathcal{N}_{i,j} = \frac{\mathcal{T}_{i,j} -  \min\limits_{(i, j) \in \mathcal{T}} \mathcal{T}_{i, j}}
	{ \max\limits_{(i, j) \in \mathcal{T}} \mathcal{T}_{i, j} - \min\limits_{(i, j) \in \mathcal{T}} \mathcal{T}_{i, j} } .
	\label{eq:Nnormalized}  
\end{equation}


We use ($ W $) as the world frame, ($ B $) as the drone body frame.
Notably, we use ($E$) to represent the event camera frame while ($D$) representing the depth camera frame.
Hence, we can represent the transformation ${^W\mathbf{T}_{E}}$ of the event camera in the world frame as 
\begin{equation}
	{^W\mathbf{T}_{E}} = \left[\begin{matrix}
		{^W\mathbf{R}_{E}} & {^W\mathbf{t}_{E}} \\
		\mathbf{0}^\top & 1 \\
	\end{matrix}
		\right]  .
\end{equation}
.

%

\section{Fast-moving Object Detection and Tracking}
\label{sec:Methedology}

\subsection{Ego-motion Compensation}
\label{Meth:EgoMotionCompensation}

\begin{figure}
	\centering
	\includegraphics[width=1.0\linewidth]{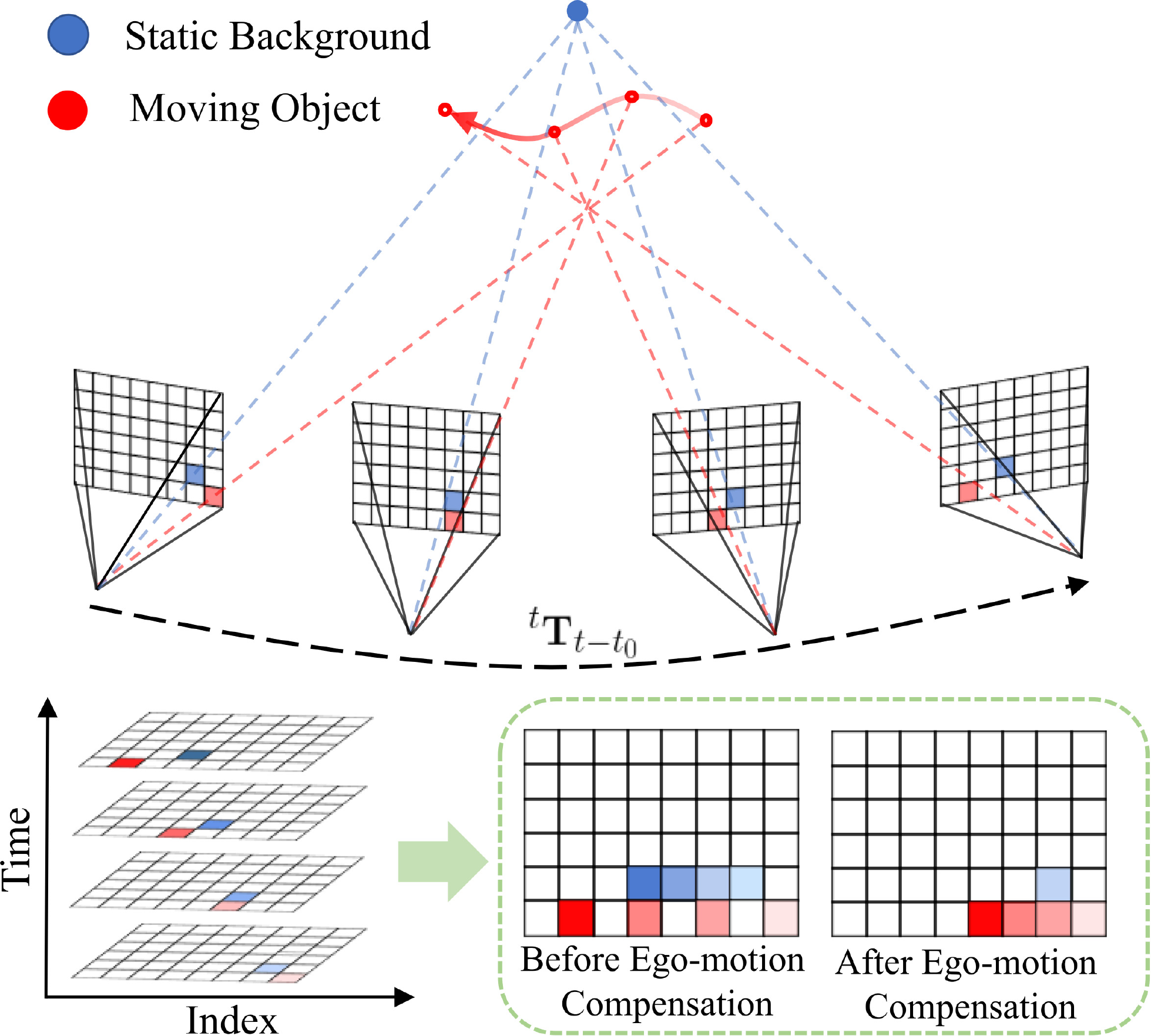}
	\caption{
	Working principle of our ego-motion compensation algorithm. Both rotation and translation of the event camera are considered in $^t\mathbf{T}_{t-t_{0}}$. Static background (blue) is projected to the same pixel and has a relatively low average timestamp. A moving object (red) cannot be compensated and has a wide range of timestamps. So they can be distinguished in terms of their different temporal distribution.
	}
	\label{fig:MoCapPipeline}
\end{figure}

Events can be triggered either by moving objects or by the ego-motion of the camera. 
In order to segment objects, events generated by ego-motion (backgrounds) should be filtered out first.
In previous works \cite{falanga2020dynamic}\cite{mitrokhin2018eventbased} \cite{mitrokhin2019evimo}, algorithms for ego-motion compensation are either computational demanding or not accurate enough.
To eliminate this problem, we present a method considering computational efficiency, accuracy, and robustness by fusing depth and IMU data to compensate for rotational and translational ego-motion. 
The illustration of this section can be inferred in Fig. \ref{fig:MoCapPipeline}.

This section will introduce our advanced ego-motion compensation method in two steps: rotational and translational. 
Before projection, we store some events into an event buffer $\mathbf{B}_{t_0}$ in a small time window $\Delta t = 25\text{ms}$ from timestamp $ t_0 $. 
Then we utilize IMU data to compute the average angular velocity $\bar{\omega}$ and orientation matrix ${^W\mathbf{R}_{C}}$ in the world frame during $\Delta t$. 


\subsubsection{Rotational Compensation}

We apply this compensation step to eliminate events generated by the camera's rotation \cite{falanga2020dynamic}. 
After getting the average angular velocity $\bar{\omega}$, we apply the Rodrigues' Rotation Formula \cite{stanley1978quaternion} to build the rotation matrix  $\mathbf{R}_{e}$ from relative angle $\bar{\omega} (t - t_0)$ at timestamp $t$. 
Instead of building this matrix at each timestamp $t$, we update it every millisecond to decrease computational costs.
Then, we use this rotation matrix $\mathbf{R}_{e}$ to apply a warp field $\phi(\omega, t_0): (x, y, t) \rightarrow (x_{rot}, y_{rot}, t_0) $ for every event on the image plane. \cite{mitrokhin2018eventbased} \cite{falanga2020dynamic}.
This event warping process can be denoted as follows 
\begin{equation}
\begin{aligned}
	C^\prime & = \phi(C) \\
				&= \phi(x, y, t - t_0) \\
				& = (x_{rot}, y_{rot}, t_0) , ~~~ \forall (x,y,t)\in C \label{con:compensation} ,
\end{aligned}
\end{equation}
with $C^\prime $ being the motion-compensated events in buffer $\mathbf{B}_{t_0} $. 
After compensation, we project compensated events to a 2D image plane by the event camera's intrinsic matrix. 

\subsubsection{Translational Compensation}

We apply this step to eliminate noise generated by the camera's translation.

Previous compensation methods \cite{falanga2020dynamic} are limited by lacking depth estimation. 
Without unreliable depth for each pixel, they cannot compensate for ego-translation, which leads to misdetection when drones are flying fast.
We solve this problem by leveraging an onboard depth camera.

We now project events on 2D camera plane $(c)$ to 3D body frame $(b)$ by perspective projection model with homogeneous coordinates \cite{HartleyRichard2004MVGi}
\begin{equation}
d \left[ \begin{array}{c}
	  i  \\
	  j  \\
	  1  \\
  \end{array} \right]  
	= \mathbf{K_{E}} \mathbf{X_E} ,
\end{equation}
where $\mathbf{X}_E$ is the event's position in the camera frame, $(i, j)$ represents event's coordinate on the image plane, $\mathbf{K_{E}}$ is the intrinsic matrix.
We then project this point into world frame by transform matrix ${^W\mathbf{T}_{E}}$ and apply this transitional compensation by multiplying matrix $\mathbf{T}_{t - t_0}$ as re-project it back to the camera frame, 
\begin{equation}
	\mathbf{X}_E^\prime ={ ^W\mathbf{T}_{E}^{-1} }{\mathbf{T}_{t - t_0}} {^W\mathbf{T}_{E}} \mathbf{X}_E
\end{equation} 
where $\mathbf{X}_E^\prime$ is event's compensated position in the camera frame. 

This translational compensation matrix $\mathbf{T}_{t - t_0}$ is built from the derivative of position over time $t - t_0$
 \begin{equation}
 	{\mathbf{T}_{t - t_0}} = \left[\begin{matrix}
 		\mathbf{I} & {\mathbf{v} \cdot (t - t_0)} \\
 		\mathbf{0}^\top & 1 \\
 	\end{matrix}
 	\right]  ,
 \end{equation}
with $\mathbf{v}$ being the velocity using estimation from our odometry. 
Due to computational cost, we update velocity $\mathbf{v}$ and timestamp $t$ to build this matrix every millisecond.


Results of our advanced ego-motion compensation algorithm can be seen in Fig. \ref{fig:egoCpr}.
After motion compensation, we can filter out the background by simply thresholding on the normalized mean-time graph $\mathcal{N}$ (see Equation \ref{eq:Nnormalized})

\subsection{Object Detection}
\label{Meth:ObjDetection}

\subsubsection{Dynamic Obstacle Segmentation}
\label{Meth:OD:DynObsSeg}

After ego-motion compensation, we propose a thresholding method to filter out the background in normalized mean-time image $\mathcal{N}$.
Instead of using a fixed threshold, we design an adaptive one considering angular and linear velocity as $ \theta(\omega, \upsilon) = \text{mean}(\mathcal{N}) + a||\omega|| + b||\upsilon|| + c$, 
where $\omega$ and $\upsilon$ is the magnitude of angular and linear velocity,  $ a $, $ b $ and $  c $ are parameters.
We use this threshold $\theta(\omega, \upsilon)$ to classify objects and background. Let us define $\mathcal{S}$ to represent the image after motion segmentation, which is formulated as: 
\begin{equation}
    \mathcal{S}_{i, j} = 
	\left\{ \begin{array}{l l}
			\mathcal{N}_{i, j}, \quad \mathcal{N}_{i, j} \geq \theta(\omega, \upsilon) \\
			\;\; 0 \;\;\, , \quad  \mathcal{N}_{i, j} \le \theta(\omega, \upsilon)
		\end{array}  \right. .
\end{equation}

Compared to the previous approach \cite{falanga2020dynamic}, our method can preserve more information of the moving object to decrease the possibility of missed detection.

\subsubsection{Iterative Gaussian fitting}
\label{Meth:OD:IterGssFt}

After dynamic obstacle segmentation, the image is composed of moving objects and background noise. 
Commonly, the patterns consist of the moving object have a relatively high mean-timestamp. 
While some pre-processes are still to be done to make the fitting effects better. 
First, mean filter and morphological operations are used to eliminate salt-and-pepper noise. 
Next, an element-wise square is ensued to enhance the image contrast further. 
After all pre-processes above, the \textbf{Algorithm} \ref{alg:IterativeGaussianFitting} is proposed to extract the moving object. 
Initially, $\mathcal{S}(\mathbf{C}_P^0, \mathbf{L}^0)$ is the origin ROI, where $\mathbf{C}_P^0 = (x, y)$ denotes the center point of ROI and $\mathbf{L}^0 = (w, h)$ denotes all initial side lengths. 
In this work, $\mathbf{C}_P^0 = (x, y)$ is pointed as the pattern with the highest mean-timestamp, $w$ and $h$ are distributed as $1/2$ of the image width and height.
 After that, the optimal $\mathbf{C}_P^*$ and $\mathbf{L}^*$ are computed through an iterative Gaussian fitting process. 
 Finally, the origin ROI $\mathcal{S}(\mathbf{C}_P^0, \mathbf{L}^0)$is converged to optimal, denoted as $\mathcal{S}(\mathbf{C}_P^*, \mathbf{L}^*)$.

\begin{algorithm}
	\caption{Iterative Gaussian Fitting}
	\label{alg:IterativeGaussianFitting}
	\KwIn{$\mathcal{S}(\mathbf{C}_P^0, \mathbf{L}^0), K \in\mathbb{Z}_+, \delta_C>0, \mathbf{\delta_L}>0$}
	\KwOut{$\mathcal{S}(\mathbf{C}_P^*, \mathbf{L}^*)$}
	\Begin
	{
		\While{$k<K$}
		{
			$\mathcal{S}(\mathbf{C}_P^k, \mathbf{L}^k) \sim \mathcal{N}(\mu,\,\sigma^{2})$\;
			$\mathbf{C}_P^{k+1} \leftarrow \mu$\;
			$\mathbf{L}^{k+1} \leftarrow 4\sigma$\;
			$J_{c} \leftarrow J(\mathbf{D}_P^{k+1}, \mathbf{T}^{k+1})$\;
			\If{ $||\mathbf{C}_P^{k+1}-\mathbf{C}_P^k|| < \delta_C \quad \textbf{and} $ \quad \\ \quad $\abs{\mathbf{L}^{k+1} - \mathbf{L}^k} < \mathbf{\delta_L}$ }{\textbf{break}}
			$k \leftarrow k+1$\;
		}
		$\mathbf{C}_P^* \leftarrow \mathbf{C}_P^{k}, \mathbf{L}^* \leftarrow \mathbf{L}^{k}$\;
		
		\Return{$\mathcal{S}(\mathbf{C}_P^*, \mathbf{L}^*)$}\;
	}
\end{algorithm}

\subsubsection{Moving Region Retrieval}
\label{Meth:OD:CntChk}

Mostly, the contour of the moving object can be extracted accurately and completely. 
While in some cases, especially when the scale of the moving object is too small on the image plane, the ROI can fail to converge.
Hence, we seek connected components to find the region that is most likely to be the moving object in a fail-converged ROI. 
After this operation, the moving object is extracted accurately in the majority of cases.

\subsection{Object Trajectory Estimation}
\label{Meth:ObjTrajEst}

\begin{figure}
	\centering
	\includegraphics[width=1.0\linewidth]{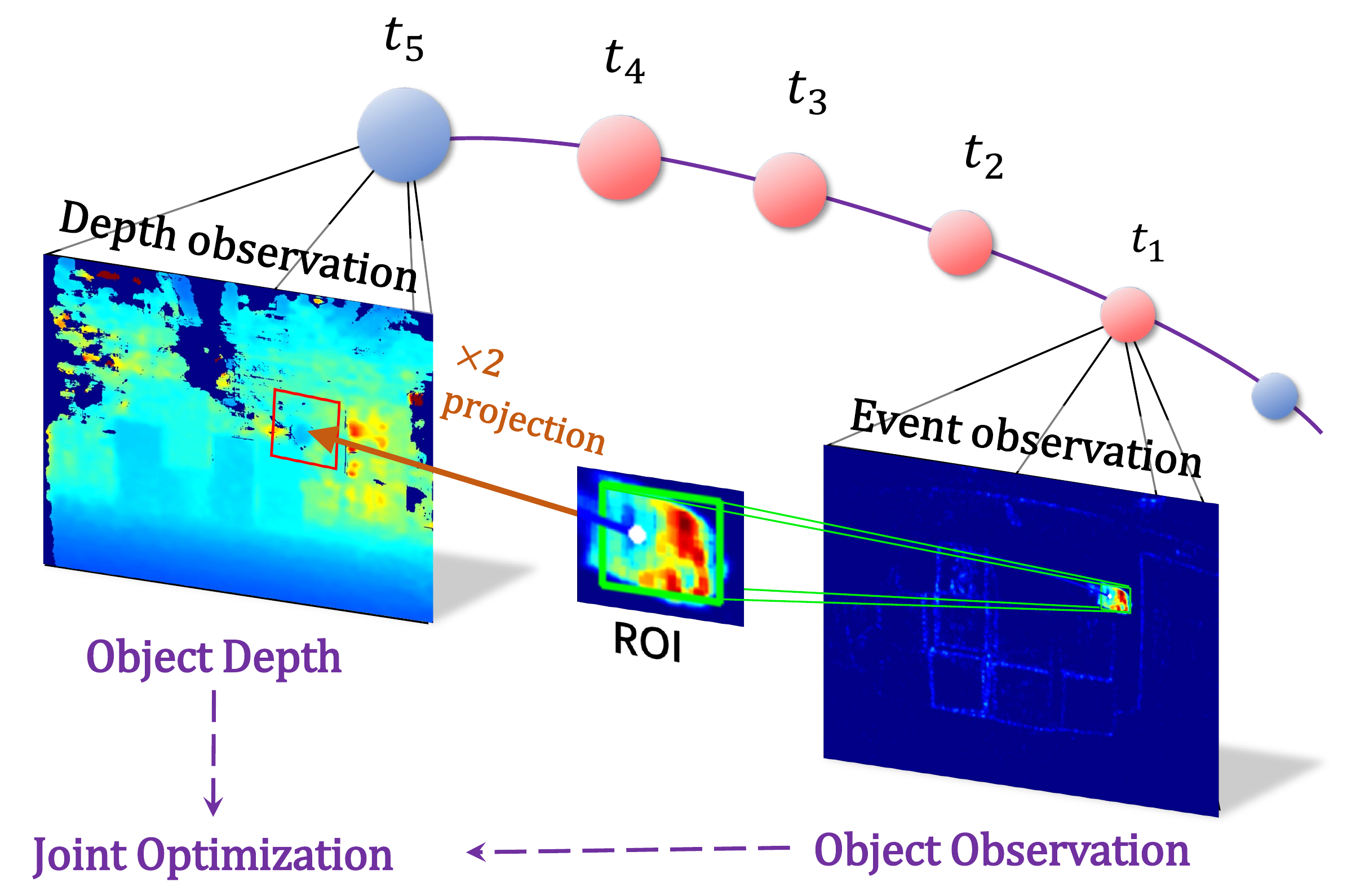}
	\caption{The pipeline of our optimization-based trajectory estimator. The pink ball and the cyan ball are the observation of event and depth separately. We use the ROI in the previous detection step (green box) and project it to the depth plane (red). Then we extract depth from the red area and location from the green area. After extracting all information, the trajectory is estimated by joint optimization.}
	\label{fig:JointOptimisation}
\end{figure}

As the moving objects are detected, a 3D trajectory can be predicted.
However, the two sensors' sample frequency is different, therefore synchronization can be challenging. 
We are inspired by \cite{su2017catching} and propose an optimization-based trajectory estimation and prediction system fusing event and depth observation without synchronous sampling.

An illustration of this section can be seen in Fig. \ref{fig:JointOptimisation}. 

This section introduces our entire system in three parts: 
(1) 2D object correspondence and tracking. 
(2) Object segmentation in the depth plane
(3) 3D trajectory estimator fusing event and depth observation.

\subsubsection{2D Object Correspondence and Tracking}
\label{Meth:OTE:2DObjTrkg}

To estimate the object trajectory from detection results, we need to know which object the result corresponds to.
When a new detection comes up, the algorithm will judge the time discrepancy and position deviation from the previous detection. 
After the correspondence is determined, the object is tracked by an Extended Kalman Filter (EKF) \cite{EKF} with a linear constant acceleration model on the 2D camera plane.
The EKF updates the object's central position on the 2D image plane to estimate its 3D position, velocity, and acceleration.

There are two reasons why we use this Extended Kalman Filter. 
Firstly, due to the event camera's limitation, the appearance of objects on the event plane is not stable and accurate, resulting in the noise of objects' 2D positions under event observation. EKF can filter this noise.
Second, potential misdetection may occur, and EKF can predict the object position in this case.

\subsubsection{Object Segmentation in the Depth Plane}
\label{Meth:OTE:ObjSegDpt}

A 3D trajectory is required for motion planning, but the tracking method only provides 2D position. 
Therefore, the perception of object depth is essential. 
We solve this by Semi-tight coupling depth segmentation. 
In other words, we use the detection results of the event camera to assist depth camera segmentation to decrease the computation and processing latency. 
The principles of our whole process are described below.

First of all, the depth map from the depth camera is registered to the event camera according to the intrinsic and extrinsic matrices of the event and depth cameras.

In practice, because the data frequency of depth and event camera is not equal, object position after projection may have a little bias. 
We scale the bounding box twice as the ROI area. 


\begin{figure}
	\centering
	\includegraphics[width=1.0\linewidth]{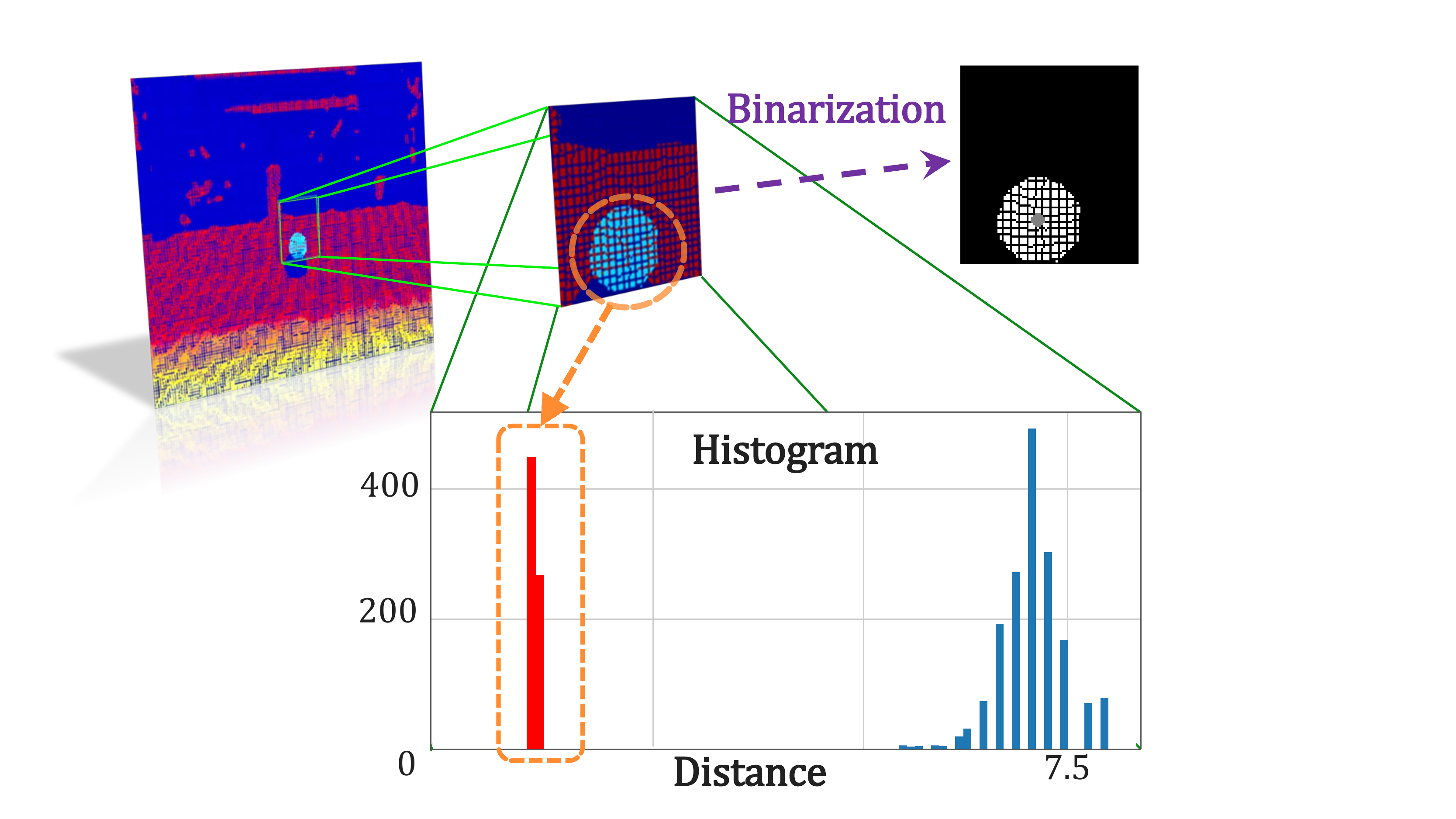}
	\caption{The process of our object depth segmentation. The minimum distance in histogram is the depth of the object due to the obstacle is the forefront in most case. The mask is created by binarization for computing the variance.}
	\label{fig:hist}
\end{figure}

After previous steps, the approximate location of the object has been determined. We assume that the most dangerous obstacle is closest to us. So the object can be separated by the nearest peak in the histogram of the depth map. The process is shown as Fig. \ref{fig:hist}.
 
To improve the system robustness,  we compute the mean and variance of segmented depth pixels. 
If the variance is too high, it might mean that these pixels belong to the background, which should not be considered. 
Otherwise, these pixels belong to the object, and we average the value to represent the camera and the object's distance.



\subsubsection{3D Trajectory Estimator Fusing Event and Depth}

Although the 2D position of the object and the depth have been estimated separately, the event camera is faster than the depth camera, so we can not associate the object depth and the 2D position on the camera plane directly. 
Inspired by \cite{su2017catching}, a 3D optimization-based trajectory estimator fusing 2D position and depth is proposed (see Fig. \ref{fig:JointOptimisation}).
The most significant difference is that we fuse the depth residual into the optimization framework. 
Before the start, two assumptions should be stated. 
First, the drone has known the earth's gravity. 
Second, the object is in free fall, ignoring air resistance. 
We describe the trajectory as Equation \ref{con:obj_traj}. 
Given the initial 3D position $p_{t_0}$ ,velocity $v_{t_0}$, gravity vector ${^Wg}$ and the start time $t_0$, we can predict the object 3D position at any time $t$ expressed as $p(t)$. 
From object correspondence, the time $t_0$ when the object first appeared can be measured. 
We just need to obtain the initial 3D position $p_{t_0}$ and velocity $v_{t_0}$ to represent the whole trajectory.

\begin{equation}
	p(t)= {^Wp_{t_0} }+ (t-t_0)  {^Wv_{t_0}} + \frac{1}{2} {^Wg } (t-t_0)^2  .
	\label{con:obj_traj}
\end{equation}

The $p_{t_0}$ and $v_{t_0}$ are estimated through the nonlinear optimization by minimizing the depth residual and reprojection error of event observation. 
Due to depth residual, the number of observations in the same period has increased so that the convergence speed is faster and the robustness of the system is improved significantly compared to the monocular method. 

At time $t_k$, we detect and track the object in the event camera and the predicted object's position in camera frame can be written as Equation \ref{con:obj_pred}. $^E p_t$ means position of object in camera frame at time $t_k$. $^W \mathbf{R}_{E_{t}}$ and $^W \mathbf{t}_{E_{t}}$ respectively represent the rotation matrix and translation vector from world to event camera. $u_t$ and $v_t$ are the object position on event camera plane from 2D tracking.
The residual $r_E$ can be written as Equation \ref{con:obj_pred} and \ref{con:event_residual}.
Meanwhile, in Equation \ref{con:event_residual}, we assume the camera model is pinhole, but this model can be changed according to the actual lens. 

\begin{equation}
	{^E p_{t_k}} = 
	\begin{bmatrix}
		x_{t_k}  \\ 
		y_{t_k}  \\  
		z_{t_k} \\
	\end{bmatrix}  = {^W \mathbf{R}_{E_{t_k}}}(p(t
	_k)- {^W \mathbf{t}_{E_{t_k}}}) ,
	\label{con:obj_pred}
\end{equation}

\begin{equation}
	r_{E} = 
	\begin{bmatrix}
		\frac{x_{t_k}}{z_{t_k}} - u_{t_k}  \\ 
		\frac{y_{t_k}}{z_{t_k}} - v_{t_k}   \\  
	\end{bmatrix}  ,
	\label{con:event_residual}
\end{equation}

Similarly, the depth residual $r_D$ is expressed as Equation \ref{con:depth_obj_pred} and \ref{con:depth_residual} with the rotation $^W \mathbf{R}_{D_{t_i}}$ and translation $^W \mathbf{t}_{D_{t_i}}$ from world frame to depth camera frame at time ${t_i}$. 
$d_{t_i}$ is the depth from depth camera observation at time $t_i$.  It should be indicated that the $r_E$ and $r_D$ are independent and various.
\begin{equation}
	{^D p_{t_i}} = 
	\begin{bmatrix}
		x_{t_i}  \\ 
		y_{t_i}  \\  
		z_{t_i} \\
	\end{bmatrix}  = {^W \mathbf{R}_{E_{t_i}}}(p(t
	_i)- {^W \mathbf{t}_{E_{t_i}}}) ,
	\label{con:depth_obj_pred}
\end{equation}

\begin{equation}
	r_{D} = 	z_{t_i} - d_{t_i}  ,
	\label{con:depth_residual}
\end{equation}

\begin{equation}
	 \min_{^Wp_{t_0},^Wv_{t_0}} \sum_{k=0}^{N} ||r_{E}||^2 + \sum_{i=0}^{M} ||r_{D}||^2
	\label{con:min_func}
\end{equation}

Then this problem can formulate nonlinear optimization problem as Equation \ref{con:min_func} to obtain the trajectory parameter $(^Wp_{t_0} ,^Wv_{t_0})$. For better robustness, we use the Huber loss.



\begin{figure}[b]
	\centering
	\includegraphics[width=1.0\linewidth]{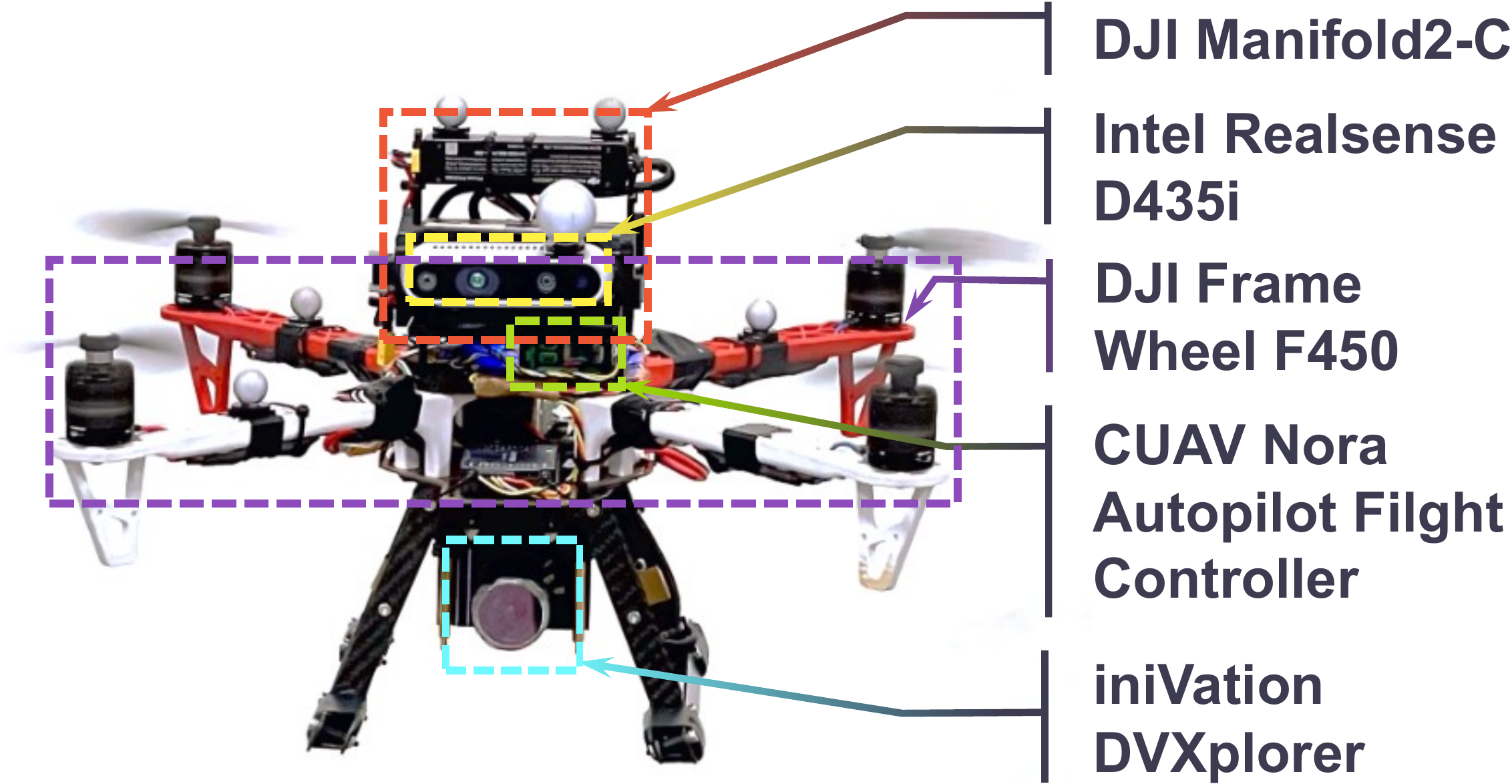}
	\caption{Overview of our UAV system}
	\label{fig:QuadSys}
\end{figure}
\section{Experiment and Evaluation}
\label{sec:Results}
\subsection{Implementation Details}
\label{Rst:IplDtls}

\begin{figure*}[b]
	\centering
	\subfigure[RGB]{
		\includegraphics[width=0.23\textwidth]{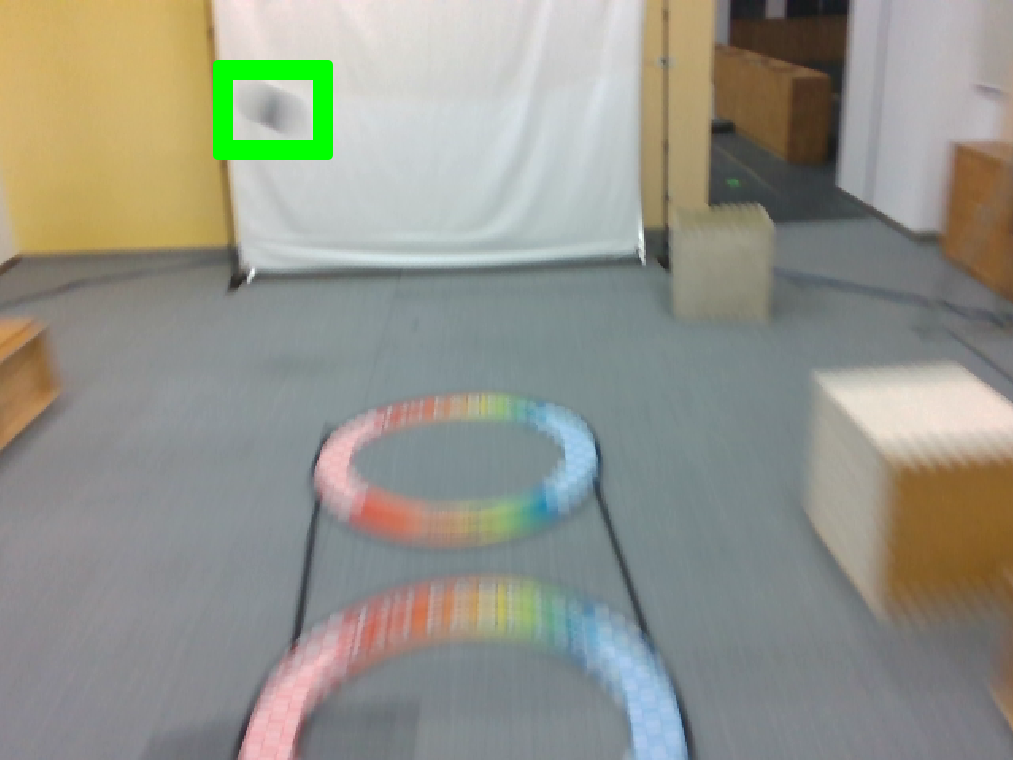}
		\label{fig:egoCprColor}
	}
	\subfigure[BetterFlow\cite{mitrokhin2018eventbased}]{
		\includegraphics[width=0.23\textwidth]{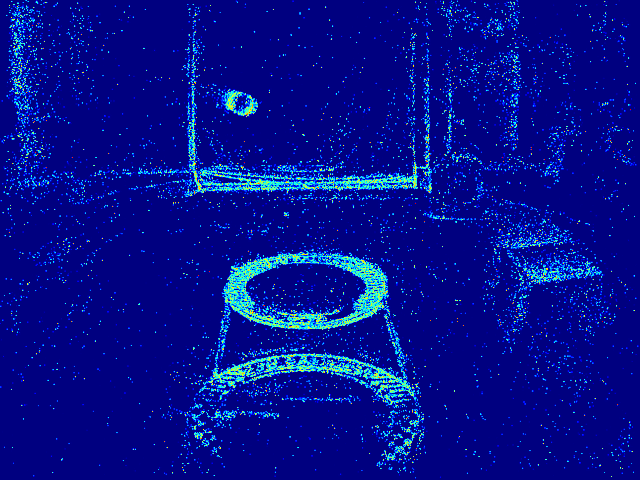}
		\label{fig:egoCprBF}
	}
	\subfigure[Falanga\cite{falanga2020dynamic}]{
		\includegraphics[width=0.23\textwidth]{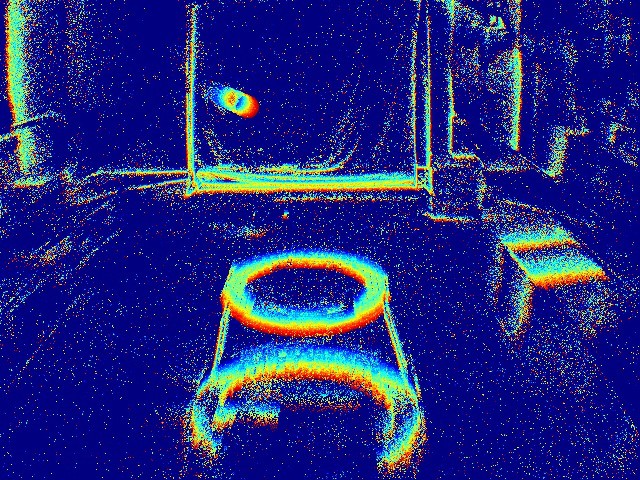}
		\label{fig:egoCprIMU}
	}
	\subfigure[ours]{
		\includegraphics[width=0.23\textwidth]{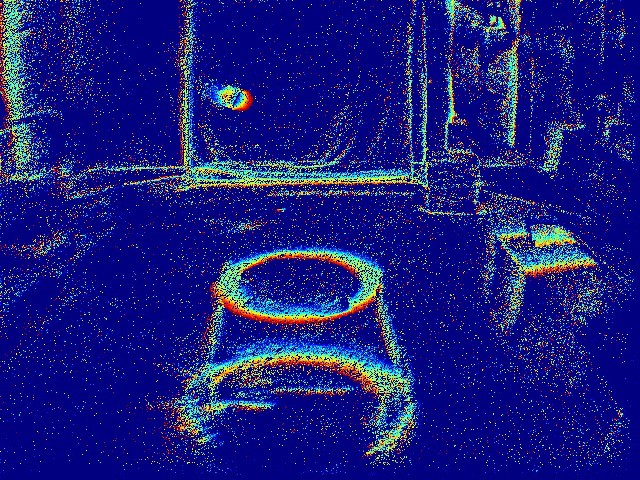}
		\label{fig:egoCprIMUD}
	}
	\subfigure[origin]{
		\includegraphics[width=0.23\textwidth]{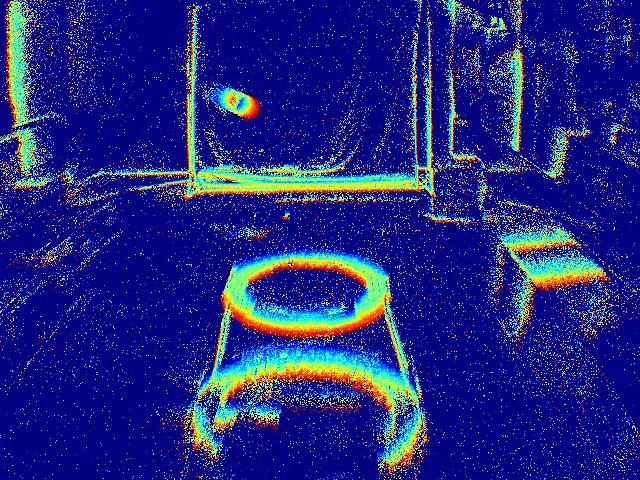}
		\label{fig:egoCprOri}
	}
	\subfigure[BetterFlow\cite{mitrokhin2018eventbased} after denoising]{
		\includegraphics[width=0.23\textwidth]{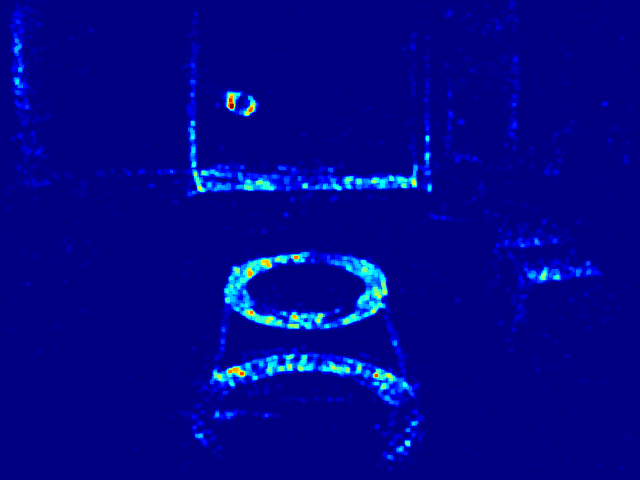}
		\label{fig:egoCprBFM}
	}
	\subfigure[Falanga\cite{falanga2020dynamic} after denoising]{
		\includegraphics[width=0.23\textwidth]{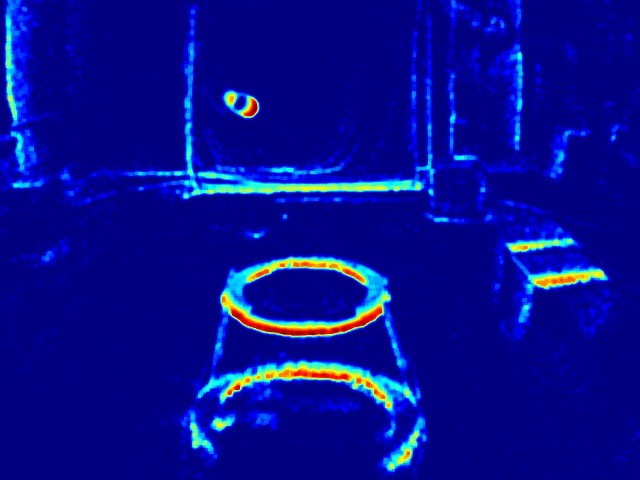}
		\label{fig:egoCprIMUM}
	}
	\subfigure[ours after denoising]{
		\includegraphics[width=0.23\textwidth]{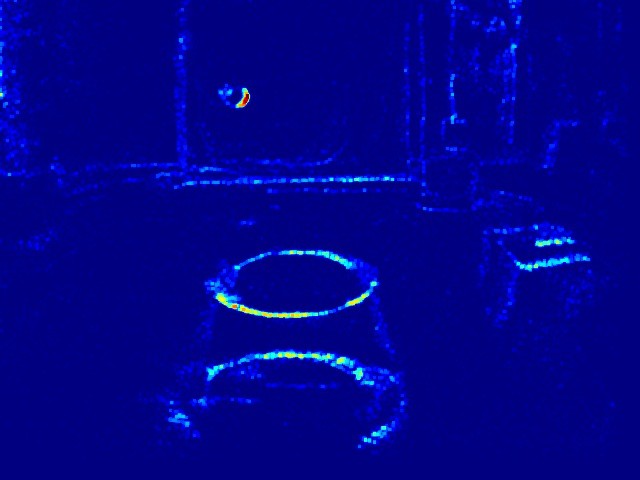}
		\label{fig:egoCprIMUDM}
	}
	\caption{
		Comparison of our ego-motion compensation method with \cite{mitrokhin2018eventbased} and \cite{falanga2020dynamic} (Section \ref{Meth:EgoMotionCompensation}). 
		Fig. \ref{fig:egoCprColor} is RGB image, which contains a flying ball (green) and two colorful banners;
		 \ref{fig:egoCprOri} is the original mean-time graph $\mathcal{T}$;
		 \ref{fig:egoCprBF} is the mean-time graph $\mathcal{T}$ generated by \cite{mitrokhin2018eventbased}, 
		 \ref{fig:egoCprIMU} is the performance using algorithm from \cite{falanga2020dynamic}; 
		 \ref{fig:egoCprIMUD} is our ego-motion method fusing IMU and depth data to compensate rotational and translational motion.
		 \ref{fig:egoCprBFM}  \ref{fig:egoCprIMUM} \ref{fig:egoCprIMUDM} are all three compensated images after unified denosing.
		 }
	\label{fig:egoCpr}
\end{figure*}

We present our real-world experiment (see section \ref{Rst:Experiment}) on a modified flight platform, carrying an iniVation DVXplorer dynamic vision sensor and an Intel Realsense D435i depth camera. 
A DJI Manifold2-C computer running Ubuntu 16.04 is mounted in our UAV for computational supports. 
We use a CUAV Nora Autopilot Flight Controller running the PX4 flight stack. 
To alleviate disturbance from the motion capture system's infrared light on the dynamic vision sensor, we add an infrared filter on the lens surface of the DVXplorer camera.
The overall weight (including LiPo battery and propellers) is 1.99 kg, with dimensions being $570 \times 570 \times 270$ mm. 
An overview of our flight platform can be seen in Fig. \ref{fig:QuadSys}.

\subsection{Evaluation of Ego-motion Compensation}

To demonstrate the robustness of our method, we put our system into high dynamic scenarios, where the UAV flies at a speed of over 5 m/s. Three algorithms are applied in two scenes, one with no moving object and another has an object that moves over 10 m/s (see TABLE \ref{tab:EgoMotCpr2}).
To ensure efficiency and accuracy, we hope the process has lower time consumption. 
Moreover, the contrast between the moving object and the background is deemed to be as high as possible, which is critical for detection algorithms. 
We call this the Relative Contrast.
To derive this, we define the manually marked bounding box of the moving object in a motion-compensated image as $\mathcal{M}$ (marked as green in Fig. \ref{fig:egoCprColor} on the RGB frame).
We depict the rest part of the image as $\mathcal{B}$. 
Then, relative contrast $\mathcal{\eta}$ is defined as:

\begin{equation}
	\mathcal{\eta} = \frac{\max\limits_{(i, j) \in \mathcal{M}} \mathcal{M}_{i, j} - \max\limits_{(i, j) \in \mathcal{B}} \mathcal{B}_{i, j}}{\max\limits_{(i, j) \in \mathcal{M}} \mathcal{M}_{i, j}} .
	\label{con:contrast_evaluation}
\end{equation}


Notice, since the sensors are imperfect, noise is introduced. 
It is meaningless to compute the relative contrast on an image that has much noise because noise often has the timestamp from oldest to newest, so they are more likely to be selected for computation of relative contrast instead of the moving object or background. 
Therefore, we apply unified denoising for images after ego-motion compensation by all three methods before computing the relative contrast. Besides, the relative contrast can only be computed in Scene 2 because scene 1 does not have moving objects.

Usually, these indicators cannot be met at the same time, so the trade-off between performance and efficiency is indispensable. 
In this work, we sacrifice a little efficiency under the promise of real-time. 
Table \ref{tab:EgoMotCpr2} lies the results of the comparison of our method against \cite{falanga2020dynamic} and \cite{mitrokhin2018eventbased}. 
The table indicates that our method largely outperforms \cite{mitrokhin2018eventbased} while lower than \cite{falanga2020dynamic} in several million-seconds. 
At the same time, the mean value and variance of our output image are closer to the optimization-based method \cite{mitrokhin2018eventbased} than \cite{falanga2020dynamic}'s.
Moreover, our method has the highest contrast between the moving object and the background, which provides convenience for object detection.

\begin{table*}
\centering
\caption{Comparison of Ego-motion Compensation Algorithms }
\label{tab:EgoMotCpr2}
\begin{tabular}{cccccccccc}
\toprule
\multirow{2}{*}{Experiment} &
  \multirow{2}{*}{Algorithm} &
  \multicolumn{3}{c}{Time (ms)} &
  \multicolumn{3}{c}{\begin{tabular}[c]{@{}c@{}}Relative \\ Contrast (\%)\end{tabular}} &
  \multirow{2}{*}{Real-time} &
  \multirow{2}{*}{Dynamic-flight} \\
  \cmidrule(lr){3-5} \cmidrule(lr){6-8}
    &
   &
  min &
  avg &
  max &
  min &
  avg &
  max &
   &
   \\ 
\midrule
\multirow{3}{*}{Scene 1} &
  BetterFlow \cite{mitrokhin2018eventbased} &
  1173.2 &
  10620.4 &
  32461.9 &
  - &
  - &
  - &
 No&
  Yes \\
 &
  Falanga \cite{falanga2020dynamic} &
  \textbf{2.7} &
  \textbf{7.1} &
  \textbf{20.6} &
  - &
  - &
  - &  Yes & No
 \\
 &
  ours &
  4.1 &
  12.9 &
  22.4 &
  - &
  - &
  - &
  \textbf{Yes} &
  \textbf{Yes}\\ \hline
\multirow{3}{*}{Scene 2} &
  BetterFlow \cite{mitrokhin2018eventbased} &
  554.2 &
  16587.6 &
  110559.0 &
  17.7 &
  27.8 &
  36.1 &
 No &
Yes \\
 &
  Falanga \cite{falanga2020dynamic} &
  \textbf{1.4} &
  \textbf{4.6} &
  \textbf{16.4} &
  -1.2 &
  1.7 &
  3.9 &
  Yes & No
 \\
 &
  ours &
  4.3 &
  8.9 &
  20.7 &
  \textbf{22.4} &
  \textbf{32.9} &
  \textbf{51.4} &
  \textbf{Yes} &
  \textbf{Yes} \\ \bottomrule
\end{tabular}
\end{table*}

\subsection{Evaluation of Trajectory Estimation}
To validate the accuracy of estimated trajectories, we compare our fusing method with the monocular method\cite{su2017catching} in the same scenarios.
Our ground truth is provided by a Vicon motion capture system.
We perform two estimation algorithms in two different scenarios, one with the drone flying fast forward and the other with the drone swinging forward.
The drone flies at 2 m/s, and a ball is thrown at about 12 m/s from one side to another in both scenes. The modules of detection and data association are fully consistent. 
Due to the fast motion of the ball, there are seven detections on the event camera and three segmentations of the depth map in 0.18 seconds in the forward scenario. 
The swinging scenario lasts for 0.16 seconds with six detections on event and three on depth. 

\begin{figure*}
	\centering
	\subfigure[Trajectory in Fast Forward]{
		\includegraphics[width=0.48\textwidth]{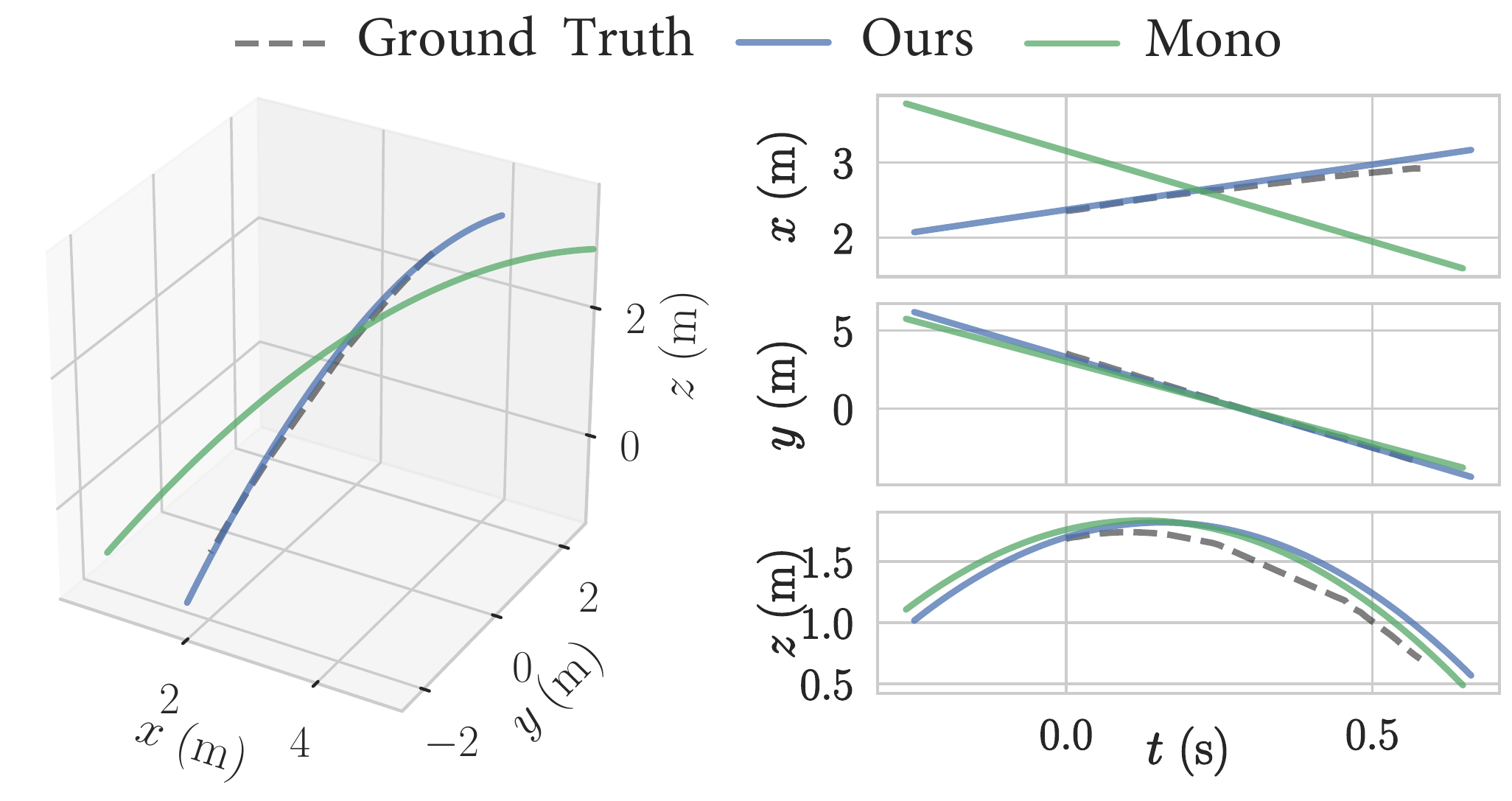}
	}
    \subfigure[Trajectory in Swing]{
		\includegraphics[width=0.48\textwidth]{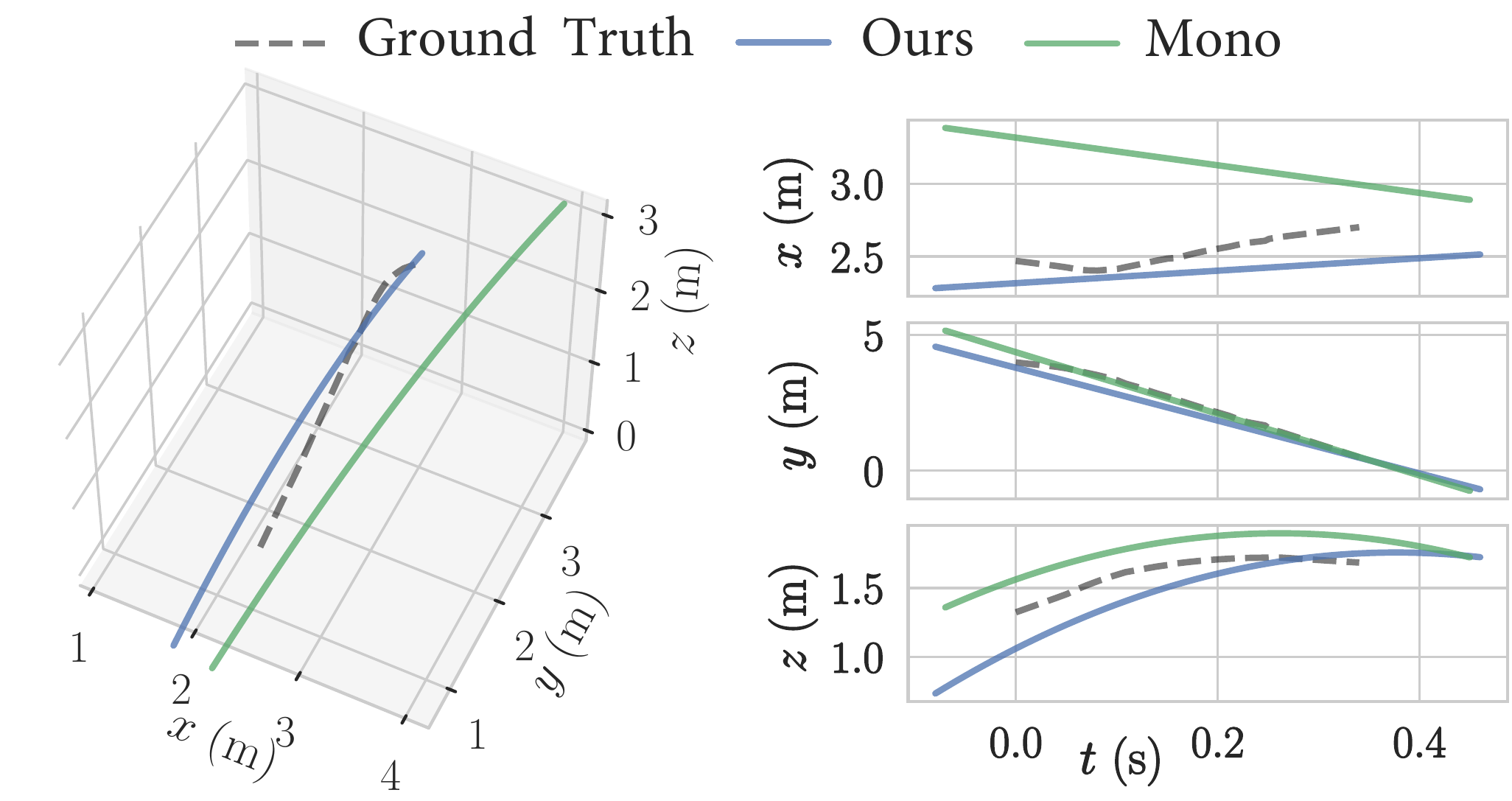}
		\label{fig:SwingTraj}
	}
	\caption{The estimated trajectories of our fusing method with the monocular method in \cite{su2017catching}. The dotted line is the ball position obtained by motion capture, the green line is the method in \cite{su2017catching} and the blue line is our fusing method. }
	\label{fig:TrajEst}
\end{figure*}

Fig. \ref{fig:TrajEst} shows a comparison of fusing two cameras versus a monocular event camera which is configured as \cite{su2017catching}.
Fig. \ref{fig:TrajEst} states that the result of fusing two sensors is significantly superior to only monocular event intuitively. 
In both scenarios, the trajectories estimated by the monocular method are opposite in the x-direction. 
It is mainly due to fewer detection times and lack of depth truth.    
We compute the APE(Absolute Pose Error) of the estimated trajectory with the reference. 
The detailed result is shown in TABLE \ref{tab:TrajCpr}. 
This comparison demonstrates that the accuracy of our method is much higher in these fast scenarios.

\begin{table}
	\centering
	\begin{threeparttable}

	\caption{Comparison of Trajectory Estimation method}
	\label{tab:TrajCpr}
	\begin{tabular}{cccccc} 
	\toprule
	\multirow{2}{*}{Experiment}   & \multirow{2}{*}{Method} & \multicolumn{4}{c}{APE (m)   }                  \\ 
	\cmidrule{3-6}
								  &                         & Mean & Min   & Max   & RMSE   \\ 
	\midrule
	\multirow{2}{*}{Fast Forward} & Mono\cite{su2017catching}                   & 0.725                    & \textbf{0.152} & 1.218 & 0.784  \\
								  & Fusion                  & \textbf{0.194}                    & 0.169 & \textbf{0.285} & \textbf{0.201}  \\ 
	\midrule
	\multirow{2}{*}{Swing}        & Mono\cite{su2017catching}                   & 0.956                    & 0.345 & 0.957 & 0.717  \\
								  & Fusion                & \textbf{0.424}                    & \textbf{0.248} & \textbf{0.562} & \textbf{0.436}  \\
	\bottomrule
	
	\end{tabular}
	\end{threeparttable}
	\end{table}

\subsection{Real-world Experiment}
\label{Rst:Experiment}

We present several throwing-ball experiments with onboard sensors both indoor and outdoor, with larger and smaller balls, bright (240 $\sim$ 1100 lux) and dim (8 $\sim$ 10 lux), swinging and moving environments.  
The main goal of these experiments is to validate our detection, tracking, and 3D trajectory estimation system onboard highlighted in the dynamic object avoidance scene.

One experiment is to throw a ball of unknown size at a hovering UAV, which would move upward after detecting the ball to avoid the collision. 
A ball with a diameter of 21cm was thrown at a distance ranging from 8 to 10 meters at speeds from 7.0 to 12.0 m/s like Fig. \ref{fig:top}.
In this experiment, background events are triggered by in-situ vibration and rotation.
We did this experiment in several scenes to evaluate the performance under different ambient illumination levels.

Another experiment is to dodge a throwing ball while the UAV is flying forward. 
The ball was thrown at the same position and speed compared to the last experiment.
Different from that one, plenty of background events were triggered by the UAV's translational motion, making the moving ball harder to detect.
Eventually, our system successfully tackles this challenge with remarkable performance (please refer to our attached video).

\section{Conclusion}
\label{sec:Conclusion}

In this paper, we present a novel perception system for solving dynamic object avoidance problems. 
It achieves a computational-friendly while accurate motion compensation for event-based object detection.
It also presents a robust 3D trajectory estimator leveraging both event and depth data.
The system has been tested in real-world experiments to prove its advantages.

Nevertheless, there is still room for improvement in some aspects. 
Integrating avoidance algorithm based on motion planning with our perception system is one of the most promising improvements. 
In this way, a carefully generated trajectory such as \cite{zhou2020egoplanner} \cite{wang2021geometrically} could consider static and dynamic scenes, avoiding performance and flight smoothness.

\newlength{\bibitemsep}\setlength{\bibitemsep}{0.00\baselineskip}
\newlength{\bibparskip}\setlength{\bibparskip}{0pt}
\let\oldthebibliography\thebibliography
\renewcommand\thebibliography[1]{
	\oldthebibliography{#1}
	\setlength{\parskip}{\bibitemsep}
	\setlength{\itemsep}{\bibparskip}
}
\bibliography{references}

\begin{thebibliography}{10}
\providecommand{\url}[1]{#1}
\csname url@rmstyle\endcsname
\providecommand{\newblock}{\relax}
\providecommand{\bibinfo}[2]{#2}
\providecommand\BIBentrySTDinterwordspacing{\spaceskip=0pt\relax}
\providecommand\BIBentryALTinterwordstretchfactor{4}
\providecommand\BIBentryALTinterwordspacing{\spaceskip=\fontdimen2\font plus
\BIBentryALTinterwordstretchfactor\fontdimen3\font minus
  \fontdimen4\font\relax}
\providecommand\BIBforeignlanguage[2]{{%
\expandafter\ifx\csname l@#1\endcsname\relax
\typeout{** WARNING: IEEEtran.bst: No hyphenation pattern has been}%
\typeout{** loaded for the language `#1'. Using the pattern for}%
\typeout{** the default language instead.}%
\else
\language=\csname l@#1\endcsname
\fi
#2}}

\bibitem{falanga2020dynamic}
D.~Falanga, K.~Kleber, and D.~Scaramuzza, ``Dynamic obstacle avoidance for
  quadrotors with event cameras,'' \emph{Science Robotics}, vol.~5, no.~40, p.
  eaaz9712, 2020.

\bibitem{su2017catching}
K.~Su and S.~Shen, ``Catching a {{Flying Ball}} with a {{Vision}}-{{Based
  Quadrotor}},'' in \emph{2016 {{International Symposium}} on {{Experimental
  Robotics}}}, ser. Springer {{Proceedings}} in {{Advanced Robotics}},
  D.~Kulić, Y.~Nakamura, O.~Khatib, and G.~Venture, Eds.\hskip 1em plus 0.5em
  minus 0.4em\relax {Springer International Publishing}, 2017, vol.~1, pp.
  550--562.

\bibitem{gallegoEventbasedVisionSurvey2020}
G.~Gallego, T.~Delbruck, G.~M. Orchard, C.~Bartolozzi, B.~Taba, A.~Censi,
  S.~Leutenegger, A.~Davison, J.~Conradt, K.~Daniilidis, and D.~Scaramuzza,
  ``Event-based vision: A survey,'' \emph{IEEE Transactions on Pattern Analysis
  and Machine Intelligence (T-PAMI)}, pp. 1--1, 2020.

\bibitem{brosch2015eventbased}
T.~Brosch, S.~Tschechne, and H.~Neumann, ``On event-based optical flow
  detection,'' \emph{Frontiers in Neuroscience}, vol.~9, 2015.

\bibitem{zhu2017eventbased}
A.~Z. Zhu, N.~Atanasov, and K.~Daniilidis, ``Event-based feature tracking with
  probabilistic data association,'' in \emph{Proc. of the {IEEE} Intl. Conf. on
  Robot. and Autom. (ICRA)}, 2017, pp. 4465--4470.

\bibitem{Piatkowska2012Spatiotemporal}
E.~{Piatkowska}, A.~N. {Belbachir}, S.~{Schraml}, and M.~{Gelautz},
  ``Spatiotemporal multiple persons tracking using dynamic vision sensor,'' in
  \emph{2012 IEEE Computer Society Conference on Computer Vision and Pattern
  Recognition Workshops}, 2012, pp. 35--40.

\bibitem{lagorce2015asynchronous}
X.~Lagorce, C.~Meyer, S.~Ieng, D.~Filliat, and R.~Benosman, ``Asynchronous
  {{Event}}-{{Based Multikernel Algorithm}} for {{High}}-{{Speed Visual
  Features Tracking}},'' \emph{IEEE Transactions on Neural Networks and
  Learning Systems}, vol.~26, no.~8, pp. 1710--1720, Aug. 2015.

\bibitem{li2019robust}
H.~Li and L.~Shi, ``Robust event-based object tracking combining correlation
  filter and cnn representation,'' \emph{Frontiers in Neurorobotics}, vol.~13,
  p.~82, 2019.

\bibitem{brandliELiSeDEventbasedLine2016}
C.~Brandli, J.~Strubel, S.~Keller, D.~Scaramuzza, and T.~Delbruck, ``{{ELiSeD}}
  — {{An}} event-based line segment detector,'' in \emph{2016 {{Second
  International Conference}} on {{Event}}-Based {{Control}}, {{Communication}},
  and {{Signal Processing}} ({{EBCCSP}})}.\hskip 1em plus 0.5em minus
  0.4em\relax {IEEE}, 2016, pp. 1--7.

\bibitem{mitrokhin2018eventbased}
A.~Mitrokhin, C.~Fermuller, C.~Parameshwara, and Y.~Aloimonos, ``Event-{{Based
  Moving Object Detection}} and {{Tracking}},'' in \emph{Proc. of the IEEE/RSJ
  Intl. Conf. on Intell. Robots and Syst.}, 2018, pp. 1--9.

\bibitem{gallego2018unifying}
G.~Gallego, H.~Rebecq, and D.~Scaramuzza, ``A {{Unifying Contrast Maximization
  Framework}} for {{Event Cameras}}, with {{Applications}} to {{Motion}},
  {{Depth}}, and {{Optical Flow Estimation}},'' in \emph{Proc. of the {IEEE}
  Intl. Conf. on Pattern Recognition (CVPR)}.\hskip 1em plus 0.5em minus
  0.4em\relax {IEEE}, 2018, pp. 3867--3876.

\bibitem{zhou2018semi}
Y.~Zhou, G.~Gallego, H.~Rebecq, L.~Kneip, H.~Li, and D.~Scaramuzza,
  ``Semi-dense 3d reconstruction with a stereo event camera,'' in
  \emph{Proceedings of the European Conference on Computer Vision (ECCV)},
  2018, pp. 235--251.

\bibitem{zhou2021eventbased}
\BIBentryALTinterwordspacing
Y.~Zhou, G.~Gallego, and S.~Shen. Event-based {{Stereo Visual Odometry}}.
  [Online]. Available: \url{http://arxiv.org/abs/2007.15548}
\BIBentrySTDinterwordspacing

\bibitem{zhou2020event}
\BIBentryALTinterwordspacing
Y.~Zhou, G.~Gallego, X.~Lu, S.~Liu, and S.~Shen. (2020) Event-based motion
  segmentation with spatio-temporal graph cuts. [Online]. Available:
  \url{https://arxiv.org/abs/2012.08730}
\BIBentrySTDinterwordspacing

\bibitem{mitrokhin2019evimo}
A.~Mitrokhin, C.~Ye, C.~Fermüller, Y.~Aloimonos, and T.~Delbruck,
  ``{{EV}}-{{IMO}}: {{Motion Segmentation Dataset}} and {{Learning Pipeline}}
  for {{Event Cameras}},'' in \emph{Proc. of the {IEEE/RSJ} Intl. Conf. on
  Intell. Robots and Syst. (IROS)}, 2019, pp. 6105--6112.

\bibitem{stanley1978quaternion}
W.~Stanley, ``Quaternion from rotation matrix,'' \emph{Journal of Guidance and
  Control}, vol.~1, no.~3, pp. 223--224, 1978.

\bibitem{HartleyRichard2004MVGi}
R.~Hartley and A.~Zisserman, \emph{\BIBforeignlanguage{eng}{Multiple View
  Geometry in Computer Vision}}.\hskip 1em plus 0.5em minus 0.4em\relax
  Cambridge: Cambridge University Press, 2004.

\bibitem{EKF}
S.~J. {Julier} and J.~K. {Uhlmann}, ``Unscented filtering and nonlinear
  estimation,'' \emph{Proceedings of the IEEE}, vol.~92, no.~3, pp. 401--422,
  2004.

\bibitem{zhou2020egoplanner}
X.~{Zhou}, Z.~{Wang}, H.~{Ye}, C.~{Xu}, and F.~{Gao}, ``Ego-planner: An
  esdf-free gradient-based local planner for quadrotors,'' \emph{IEEE Robotics
  and Automation Letters}, vol.~6, no.~2, pp. 478--485, 2021.

\bibitem{wang2021geometrically}
\BIBentryALTinterwordspacing
Z.~Wang, X.~Zhou, C.~Xu, and F.~Gao. Geometrically {{Constrained Trajectory
  Optimization}} for {{Multicopters}}. [Online]. Available:
  \url{http://arxiv.org/abs/2103.00190}
\BIBentrySTDinterwordspacing

\end{thebibliography}

\end{document}